\documentclass[conference]{IEEEtran}
\IEEEoverridecommandlockouts
\usepackage{cite}
\usepackage{amsmath,amssymb,amsfonts}
\usepackage{algorithmic}
\usepackage{graphicx}
\usepackage{textcomp}
\usepackage{xcolor}
\usepackage{wrapfig}
\graphicspath{{./images/}}

\def\BibTeX{{\rm B\kern-.05em{\sc i\kern-.025em b}\kern-.08em
    T\kern-.1667em\lower.7ex\hbox{E}\kern-.125emX}}

\begin{document}
\title{Uplift Modeling for Multiple Treatments with Cost Optimization}

\author{
\IEEEauthorblockN{Zhenyu Zhao}
\IEEEauthorblockA{\textit{Uber Technologies, Inc.} \\
San Francisco, USA \\
zhenyuz@uber.com}
\and
\IEEEauthorblockN{Totte Harinen}
\IEEEauthorblockA{\textit{Uber Technologies, Inc.} \\
San Francisco, USA \\
totte@uber.com}
}


\maketitle

\begin{abstract}
Uplift modeling is an emerging machine learning approach for estimating the treatment effect at an individual or subgroup level. It can be used for optimizing the performance of interventions such as marketing campaigns and product designs. Uplift modeling can be used to estimate which users are likely to benefit from a treatment and then prioritize delivering or promoting the preferred experience to those users. An important but so far neglected use case for uplift modeling is an experiment with multiple treatment groups that have different costs, such as for example when different communication channels and promotion types are tested simultaneously. In this paper, we extend standard uplift models to support multiple treatment groups with different costs. We evaluate the performance of the proposed models using both synthetic and real data. We also describe a production implementation of the approach.
\end{abstract}



\section{Introduction}
Uplift modeling \cite{Guelman2015-qe, Gutierrez2016-co, Kunzel2017-ko, Rzepakowski2012-br, Soltys2015-be, Wager2015-sd, Zaniewicz2013-rt, Zhao2017-kg} is a technique to estimate and predict the individual-level or subgroup-level causal effects of different treatments in an experiment. This type of information is useful for designing and offering a personalized experience to improve user experience, satisfaction, and engagement. Uplift modeling is therefore commonly used in areas such as marketing, customer service, and product offering. 

It is helpful to think about uplift modeling in the context of randomized experiments (also known as A/B testing \cite{kohavi2013online, kohavi2009controlled, zhao2017inform}). In a typical experiment, users are randomly assigned to each treatment group and causal effects are then estimated for the population. Generally, the treatment that yields the best overall effect is selected and launched to all users. The uplift modeling approach also often starts with a randomized experiment to collect the data to train the model. However, a major difference is that the uplift model estimates the treatment effect at an individual or subgroup-level. This allows for a more optimal policy for launching the treatments: each individual can be allocated to the treatment group that the model predicts to have the optimal treatment effect for them.

As an example, a marketing manager may have multiple channels to reach out to a user (email, mail, SMS, in-app notification, customer service phone call, etc.) to provide support or information, an A/B testing can be used to estimate which channel would yield the best overall user engagement, and then this winning channel will be applied to all users. But if the marketing manager knows the preferred channel for each user, then he or she can reach out through each user's preferred channel. The uplift model would provide such estimation on the preferred experience the user would like to have to enable personalized experience.

Uplift modeling is different from standard predictive models in marketing in that it attempts to predict the \textit{treatment effect} rather than whether or not a user is likely to convert or not. These two are not necessarily the same. For example, a user might have a high probability to convert in any case, independently of whether he or she receives a marketing communication. This type of user would have a high conversion probability in a standard predictive model. However, his or her predicted \textit{uplift} would not likely be very high because his or her propensity to convert even in the absence of the marketing campaign. Uplift modeling is needed because a high treatment effect and a high probability to convert do not necessarily coincide.

The focus of this paper is to discuss and evaluate uplift modeling approaches for two common practical challenges: (1) when there are multiple treatment groups present and (2) when there are different costs associated with the treatments. The existing literature on uplift modeling has largely focused on the situation where there is just one treatment and one control. However, a large proportion of randomized experiments in the industry have multiple treatment groups. This is because industry experimenters typically have a number of options they want to test, such as different versions of a product feature, different channels for reaching out, different products to recommend, different promotions to send out, etc. 

There are a few existing uplift modeling methods proposed for multiple treatment groups (such as \cite{Rzepakowski2012-br, Zhao2017-kg}), however, there are sparse existing implementations of the state-of-the-art algorithms in the multi-treatment context to support industry applications. It also leads to little evidence as to which algorithms in fact \textit{are} state-of-the-art in the multi-arm context in practice in terms of model accuracy and computation performance. In this paper we aim to bridge this gap by extending two meta-learner models (\cite{Nie2017-uz, Kunzel2017-ko}) originally designed for one treatment and one control case to multiple treatment group case, and then providing empirical evaluation across different models. The advantages of the meta-learners are fast computation speed and easy to implement as they can leverage the existing implementation of standard classification and regression models.

Another gap we aim to bridge is the following. Most uplift modeling literature has focused on the situation in which there is \textit{one} outcome metric to optimize, without any constraints or weights on different treatments. However, this is rarely the situation in industry experiments. Consider a marketing manager who wants to identify the most suitable promotion out of many potential candidates to be sent to each user. Using uplift modeling, he or she could identify the treatment for each individual that results in, say, the highest probability of conversion. However, this information is usually \textit{not} sufficient for the marketing manager to make a decision between the treatment groups because each of the groups have different costs associated with them. To find the best policy, the marketing manager needs to optimize for conversion \textit{while} taking into account the cost of each treatment.

The cost for a treatment can happen when the treatment experience is served to users, such as when a customer agent makes a phone call or sends paper mail. But it can also happen when a certain event is triggered. For example, an email with a promotion costs almost nothing to send, but there is a cost to the company when the user redeems the promotion. In addition, the cost can be generalized to describe potential inconveniences for users. For example, sending an irrelevant notification to a user may cause them to unsubscribe from communications. Such costs should be considered in practice during optimization. Consequently, the algorithms developed below can take these different types of costs into account.

The algorithms have been implemented in a Python package as a horizontal solution for uplift modeling at Uber. In addition, we are in progress of implementing the described models in an automated on-demand machine learning platform. The necessary inputs for the uplift modeling creation on the platform include user cohort, the target variable (e.g. conversion), user features, and the randomized experiment tag for collecting the training data. The output of the platform is a trained uplift model and predicted uplift scores at a user level rendered to other services or dumped into a data file such as a HIVE table \cite{thusoo2009hive}.

The contributions of this paper include:
\begin{itemize}
    \item Extending meta-learners (\textit{X-Learner} \cite{Kunzel2017-ko} and \textit{R-Learner} \cite{Nie2017-uz}) to support the multiple treatments case
    \item Proposing a net value optimization framework by modifying the meta-learners to maximize the outcome with treatment cost considered for multiple treatments
    \item Empirically evaluating the proposed algorithms using synthetic and real datasets
    \item Describing an approach and design to implement the uplift modeling as a platform solution
\end{itemize}

The structure of the paper is as follows. In Section \ref{related_work}, we review the existing literature related to our work. Section \ref{uplift_causal_inference} briefly discusses the motivation and basic idea of uplift modeling. And then Section \ref{sec:multi_treatment} moves to the specific problem of uplift modeling with multiple treatment groups and discuss the problem of multiple treatments with different costs. Then, Section \ref{sec:proposed_model} introduces the algorithms we compare in the empirical evaluation. Section \ref{empirical_evaluation} then compares the performance of these algorithms using both synthetic dataset and real dataset from an  experiment. Section \ref{implementation} describes the system design for implementing the uplift modeling at scale as a platform solution. Finally, Section \ref{conclusion} concludes the paper with a discussion on the results.

\section{Related Work}  \label{related_work}
Early work in the uplift modeling framework (including \cite{Hansotia2001-vf,Radcliffe2007-oo,Guelman2015-qe,Rzepakowski2012-br,Zaniewicz2013-rt}) did not explicitly frame uplift modeling as a causal inference task, but the focus of these authors is using machine learning to determine the individuals in an experiment that benefit the most from a treatment. 
By contrast, there are researchers focused on making statistical inference for heterogeneous treatment effect (rather than finding the optimal treatment). 
Early proposals to use machine learning to estimate heterogeneous treatment effects include \cite{Athey2015-jd,Su2009-ly,Green2012-xf}. 
More recent work in this tradition include the contributions by \cite{Wager2015-sd,Richard_Hahn2017-kz,Chernozhukov2017-ix,Powers2018-fm,Lu2018-hb,Shalit2016-uj}. The majority of this literature doesn't make an explicit connection to the uplift modeling approach, although the task it considers is analogous. The review by Gutierrez and Gerardy \cite{Gutierrez2016-co} establishes the connection between these two areas of research.

The meta learning approach to uplift modeling, which we develop further in this paper, is based on combining standard machine learning approaches to estimate how treatment effects vary across subgroups. One of the simplest models, which we will call the ``\textit{Two Model}'' approach, is based on fitting a separate model for the control and treatment observations and combining these to estimate heterogeneous treatment effects. An early example of such an approach can be found in \cite{Hansotia2001-vf}. More recently, \cite{Kunzel2018-sn,noauthor_undated-xm,Nie2017-uz} have developed more complex and better performing meta-learning approaches.

There are a few existing papers that consider uplift modeling with multiple treatment groups. Rzepakowski and Jaroszewicz \cite{Rzepakowski2012-br} put forward decision tree based methods for uplift modeling that use one of the following splitting criteria: Kullback-Leibler (\textit{KL}) divergence, squared Euclidean distance (\textit{\textit{ED}}) and chi-squared divergence. The authors compare the \textit{KL} and \textit{\textit{ED}} based models to a few baseline approaches, including the \textit{Two Model} method, using the Area Under the Uplift Curve (discussed in detail below) as the criterion. Using datasets from the UCI repository \cite{asuncion2007uci} but with simulated treatment effects, the authors found that their proposed methods outperform the baselines in the two-treatment case. They also demonstrate the benefits of applying uplift modeling in the multiple treatments context.

Zhao and Fang \cite{Zhao2017-kg} propose the Contextual Treatment Selection (\textit{CTS}) algorithm, which extends to an arbitrary number of treatments. They compare the \textit{CTS} method against the \textit{Two Model} approach implemented with various machine learning algorithms and using a synthetic dataset with multiple treatments. The study finds that the \textit{CTS} approach outperforms the \textit{Two Model} method in the multiple treatments case. The \textit{CTS} method is also compared against the \textit{Two Model} approach and a third method implemented in \cite{Guelman2014-kv} using real datasets with a two-arm design, where \textit{CTS} algorithm shows a better result for optimization task.

It is also worth briefly mentioning the closely related field of multi-armed bandits \cite{vermorel2005multi} in which the focus is on online learning algorithms with an exploration-exploitation trade-off. The standard multi-armed bandit approach tries to identify the optimal treatment for all users irrespective of the characteristics of the users. Contextual multi-armed bandits \cite{Pavlidis2008-lf,Abbasi-yadkori2011-rs,Li2010-le,Langford2007-em, lu2010contextual} enable personalized experience by finding optimal treatment based on ``side information’’, which could include features related to the users. While online multi-armed bandits focus on the exploration-exploitation trade-off by reallocating the sample iteratively, the uplift modeling approach is based on fixed horizon experiments (standard A/B testing). This focus comes with certain benefits, such as a better statistical power and the ability to avoid certain pitfalls of  multi-armed bandits’ relocation approach, including Simpson’s paradox \cite{kohavi2017online}.

An adjacent approach is that of offline contextual bandits. The methods in this family seek to find the optimal action for a user given data of past action responses. \cite{Li2010-ur,Dudik2015-ul,Li2015-ou} This approach resembles uplift modeling because of its focus on offline data. However, as \cite{Li2018-mh} observe, the goals of the two approaches are subtly different. While offline contextual bandits seek to maximize the expected response to an \textit{action}, the uplift modeling approach seeks to maximize the expected \textit{uplift}. Finally, some recent work has sought to unify the approaches. For example, \cite{sawant2018contextual} leverages simple uplift modeling approaches (K-Nearest Neighbor and \textit{Two Model} approach \cite{Hansotia2001-vf}) in the multi-armed bandit framework.

\section{Uplift Modeling as a Causal Inference Task} \label{uplift_causal_inference}
It is helpful to understand uplift modeling as a causal inference problem. \cite{Gutierrez2016-co} Here, we are adopting the commonly used Neyman-Rubin causal model that is based on the potential outcomes framework \cite{Rubin1974-xa,Neyman1923-kb, Rubin2005-cz, Holland1986-dw}. Using this approach, we can represent some outcome of interest $Y$ for individual $i$ under treatment condition $t$ as $Y_i(t)$. Consider a two-arm trial and let $t = 0$ for the control condition and $t = 1$ for the treatment condition. Then, in the potential outcomes framework, we express the causal effect of the treatment as:
\begin{equation}
    Y_i(1) - Y_i(0)
\end{equation}

Let $X$ denote a vector of features and $x_i$ denote the feature values of an individual. Then we can express the conditional average treatment effect (CATE) as:
\begin{equation}
    \tau(x_i) = \mathbb{E}[Y_i(1) - Y_i(0) \mid X = x_i]
\end{equation}
Early discussions of this quantity can be found in \cite{Hahn1998-fe,Heckman1997-sl} and a recent review is in \cite{Abrevaya2015-mv}. The CATE is of special interest to researchers because it allows them to understand how treatment effects vary depending on the observed characteristics of the population of interest. This is, of course, exactly the information that is needed to target treatments effectively in the population. Consequently, uplift modeling can be seen as the task of using machine learning approaches to estimate the CATE. \cite{Gutierrez2016-co} 

Two major approaches to uplift modeling have emerged in the past 10 years. The first type of approach consists of "meta-learners", which are standard machine learning algorithms combined in various ways to estimate the CATE. \cite{Nie2017-uz,Kunzel2017-ko} 

The second type of approach to uplift modeling is based on modifying familiar machine learning algorithms such as classification and regression trees. \cite{Athey2016-on,Wager2015-sd,Guelman2015-qe,Guelman2012-bx,Rzepakowski2012-br,Athey2015-jd} As an example, consider the recent Causal Random Forest algorithm introduced by \cite{Wager2015-sd}. This approach modifies the splitting criterion of the random forest algorithm in such a way that the sample is partitioned based on the heterogeneity of treatment effects in the resulting subgroups (rather than a standard classification split criterion).

This paper mainly focuses on discussing and extending the meta-learner approaches, which have two main advantages. First, they are easy to implement since standard base models can be used in the framework without changing the underlying algorithm. Second, meta-learners are computationally efficient since they can leverage the existing base model implementation that have been already optimized for code efficiency. Several popular meta-learners designed for one control and one treatment case are introduced below. In the following section, we introduce the meta-learners discussed in this paper.

\section{Meta-learners for uplift modeling}

\subsection{\textit{Two Model} approach}
Perhaps the best known meta-learner is the \textit{Two Model} approach, which was introduced for the first time almost 20 years ago. \cite{Hansotia2001-vf}
The \textit{Two Model} approach estimates the CATE by
\begin{equation}
    \hat{\tau}(x_i) = \hat{\mu}_1(x_i) - \hat{\mu}_0(x_i)
\end{equation}
where $\hat{\mu}_1(x_i)$ and $\hat{\mu}_0(x_i)$ are model estimators for the conditional mean $\mu_1(x) := \mathbb{E}[Y(1) \mid X = x]$ and $\mu_0(x) := \mathbb{E}[Y(0) \mid X = x]$.

However, as our experiments below show, this simple meta-learner does not perform particularly well in many scenarios compared with other meta-learner approaches, such as \textit{X-Learner} \cite{Kunzel2017-ko} and \textit{R-Learner} \cite{Nie2017-uz}. 

\subsection{X-Learner}
The \textit{X-Learner}, which was proposed by \cite{Kunzel2017-ko}, starts by estimating the response functions $\mu_0(x)$ and $\mu_1(x)$ using any suitable regression methods and the data from the control and treatment groups, respectively. It then proceeds to estimate ``pseudo-effects'' $D_i$ for the observations in the control group as
\begin{equation}
\tilde{D}_i^0 = \hat{\mu}_1(x) - Y_i
\end{equation}
and for the individuals in the treatment groups as
\begin{equation}
\tilde{D}_i^1 = Y_i - \hat{\mu}_0(x)
\end{equation}
where $Y_i$ is the observed value for the user. 
The pseudo-effects are then used as the outcome in another pair of regression methods to obtain the response functions $\hat{\tau}_0(x)$ and $\hat{\tau}_1(x)$ for the control and treatment groups, respectively. Finally, these two estimators are combined to obtain the CATE as in \cite{Kunzel2017-ko}
\begin{equation}
\hat{\tau}(x) = \hat{e}(x)\hat{\tau}_0(x) + (1 - \hat{e}(x))\hat{\tau}_1(x)
\end{equation}
where $e(x)$ is the propensity score $\mathbb{P}[W_i = 1 \mid X_i = x]$ with $W_i$ indicating the treatment assignment. For a detailed discussion of the algorithm and some related approaches, see \cite{Kunzel2017-ko, Kunzel2018-sn}.

\subsection{R-Learner}
The other promising meta-learner is the \textit{R-Learner} proposed by \cite{Nie2017-uz}. This approach uses any suitable method to create the propensity score estimator $\hat{e}(x)$ and the mean outcome estimator $\hat{m}(x)$ for $m(x) = \mathbb{E}[Y \mid X = x]$.

CATE is then estimated as the following minimization task in \cite{Nie2017-uz}:

\begin{equation}
\label{eqn:rlearner-tau}
\begin{split}
\hat{\tau}(\cdot) = argmin_{\tau}\bigg\{ \frac{1}{n} \sum\limits_{i=1}^n \big(\big(Y_i - \hat{m}^{(-i)}(X_i)\big) \\
- \big(W_i - \hat{e}^{(-i)}(X_i)\big)\tau(X_i)\big)^2 + \Lambda_n(\tau(\cdot))\bigg\}
\end{split}
\end{equation}
where $\hat{e}^{(-i)}(X_i)$ and $\hat{m}^{(-i)}(X_i)$ denote predictions made without the $i$-th observation and $\Lambda_n(\tau(\cdot))$ is some regularizer of choice. The reader should consult \cite{Nie2017-uz} for details.

\section{Uplift Modeling for Multiple Treatment Groups with Different Costs} \label{sec:multi_treatment}

\subsection{Use Cases with Multiple Treatment Groups}
Many use cases in practice contain multiple treatment groups, since there are often more than one new proposals for designing the product feature, communication content, promotion types, etc. 

Below are two example use cases with multiple treatment groups, representing two typical use case categories:
\begin{itemize}
    \item Multiple Treatment Groups with a Control: One marketing manager wants to make a personalized campaign strategy for each user: not sending the email (control), sending the email in template 1 (treatment 1), or sending the email in template 2 (treatment 2). The optimization task is to determine whether to send the email and which email template should be used for each user. 
    \item Multiple Treatment Groups Without a Control: One product team developed a new web page (visible to everyone so there is no default control) with three modules, and the team would like to personalize the module rank order for each user. There are in total $6$ different possible rank order combinations as treatments. The optimization task is to find the best rank order for each user. 
\end{itemize}

\subsection{Use Cases with Different Costs}
The uplift modeling approaches mentioned so far can be used for optimizing the CATE based on the assumptions that each treatment has equal cost. However, in practice, there are many scenarios with heterogeneous treatment cost and the optimization task needs to put cost into consideration. 

There are two types of common treatment cost structures: (1) Fixed impression cost for each treated user. For example, sending a marketing campaign to users through paid channels (SMS, mail, ads, etc.) leads to a fixed cost for each send independent of the user's response. (2) Triggered cost for each converted user. For example, a promotion involving dollar values (e.g. $10\%$, $15\%$, $20\%$ discount) will only take effect when the user places the order, and there is no promotion cost for users who do not make a purchase. The impression cost and triggered cost can be different among treatments because the treatment is delivered through different channels or involves different promotions.

We define the net value for a user as the conversion value offsetting the treatment cost. It is often desirable to maximize the conversion value while controlling the associated cost. Therefore, optimizing the net value is useful in a variety of use cases. While the models mentioned in previous section can be applied for optimizing the conversion rate, they cannot be directly applied for optimizing net value.

\section{Proposed Models} \label{sec:proposed_model}
In this section, we describe the algorithms we propose to model uplift for multiple treatment groups as well as algorithms taking cost into account for optimization. In particular, we extend \textit{X-Learner} and \textit{R-Learner} to multiple treatment groups case. To our knowledge, neither of the algorithms have been extended to the multi-treatment context previously, so we are interested in evaluating how well they perform in this practically important setting. After explaining the two algorithms and the way in which we extend them to the multi-treatment case, we outline our strategy for incorporating information about the costs of different treatments into these uplift models.

\subsection{Meta-learners for Multiple Treatment Groups}

\subsubsection{Extending X-Learner for Multiple Treatments}
Consider an experiment with a control group and $m$ treatment groups. Here, we use any suitable regression method to estimate the response functions under each group:
\begin{equation}
\label{eqn:xlearner-mu}
\begin{split}
\mu_{t_j}(x) = \mathbb{E}[Y(t_j) \mid X = x]
\end{split}
\end{equation}
where $t_j \in \{t_0,t_1, ..., t_m\}$ with $t_0$ denoting the control group. We estimate $\mu_{t_0}(x)$ using data from the control group and $\mu_{t_j}(x)$ using data from the $j$-th treatment group. We then proceed to estimate the pseudo-effects analogously, ie
\begin{equation}
\label{eqn:xlearner-pseudo-effects}
\begin{split}
\tilde{D}_i^{t_0} = \hat{\mu}_{t_j}(x) - Y_i \\
\tilde{D}_i^{t_j} = Y_i - \hat{\mu}_{t_0}(x)
\end{split}
\end{equation}
where $\tilde{D}_i^{t_0}$ is estimated using control group data and $\tilde{D}_i^{t_j}$ is estimated using the data from the $j$-th treatment group. Finally, as in the two-group experiment, we use the pseudo-effects as the outcomes in another pair of regressions to obtain $\hat{\tau}_{t_0}(x)$ and $\hat{\tau}_{t_j}(x)$.

In the multiple treatment group case, we need to estimate the propensity score
\begin{equation}
\label{eqn:xlearner-propensity}
    e_{t_j}(x) = \mathbb{P}[W_i = t_j \mid X = x]
\end{equation}
for each $m$ experiment groups. Focusing on the case in which we compare each treatment group against the control, we then estimate the CATE for a given treatment group as follows:
\begin{equation}
\hat{\tau}^{t_j}(x) = \frac{\hat{e}_{t_j}(x)}{\hat{e}_{t_j}(x) + \hat{e}_{t_0}(x)}\hat{\tau}_{t_0}(x) + \frac{\hat{e}_{t_0}(x)}{\hat{e}_{t_j}(x) + \hat{e}_{t_0}(x)}\hat{\tau}_{t_j}(x)
\end{equation}
Finally, to predict the best treatment group for an individual, we estimate $\tau^{t_j}$ as
\begin{equation}
\label{eqn:xlearner-predict-cate}
\begin{split}
\hat{\tau}^{t_j}(X_i) = \frac{\hat{e}_{t_j}^{(- 1)}(X_i)}{\hat{e}_{t_j}^{(- 1)}(X_i) + \hat{e}_{t_0}^{(- 1)}(X_i)}\hat{\tau}_{t_0}^{(-1)}(X_i) \\
+ \frac{\hat{e}_{t_0}^{(- 1)}(X_i)}{\hat{e}_{t_j}^{(- 1)}(X_i) + \hat{e}_{t_0}^{(- 1)}(X_i)}\hat{\tau}_{t_j}^{(-1)}(X_i)
\end{split}
\end{equation}
where notation of the form $\hat{e}_{t_j}^{(-1)}$ indicates that $\hat{e}_{t_j}$ has been estimated without using the $i$-th observation. We then simply compare which treatment group gives the highest predicted uplift for the individual and recommend that as the treatment group if the costs of treatments are equal.

\subsubsection{Extending R-Learner for Multiple Treatments}
We use the same strategy for the \textit{R-Learner}. In the first step, we estimate the propensity scores $\hat{e}_{t_j}(x)$ and mean outcomes $\hat{m}_{t_j}(x)$ using a suitable regression approach for each task. We then plug these estimators in the minimization task given in Equation \ref{eqn:rlearner-tau}, which estimates the CATE of a given treatment for an individual in a holdout sample. The recommended treatment group for an individual is the one that gives the best uplift out of the $m$ treatments, if the costs of the treatments are equal.

\subsection{Net Value Optimization for Multiple Treatment Groups with Different Costs}
This section proposes modified meta-learner models that estimate and optimize the net value uplift in a multi-arm context.

The following notations are used for formulating the value and cost structure:
\begin{itemize}
    \item $v$: conversion value, assuming each conversion is weighted equally and the conversion value is a constant given as prior knowledge. 
    \item $c_{t_j}$ for $j \in \{0,1,2,..,m\}$: impression cost for each treatment paid for each treated user, for example, the channel cost per send. 
    \item $s_{t_j}$ for $j \in \{0,1,2,..,m\}$: triggered cost for each treatment paid for each converted user, for example, the promotion code applied by the user for purchase. 
\end{itemize}

The expected net value for user $i$ under treatment $t_j$ can be defined as $E[(v - s_{t_j}) Y_{t_j} - c_{t_j} \mid X = x]$. 

The corresponding net value CATE can be expressed as
\begin{eqnarray}
\label{eqn:nvcate}
    \tau^{t_j}(x, v, s_{t_j}, c_{t_j}) = & \mathbb{E}[(v - s_{t_j}) Y_{t_j} - (v - s_{t_0}) Y_{t_0} \\
    & - (c_{t_j} - c_{t_0}) \mid X = x]
\end{eqnarray}

In the following two sections, we propose modifications to two promising meta-learners to estimate the net value CATE.

\subsubsection{Extending X-Learner for Net Value CATE}
To modify the \textit{X-Learner} for estimating the net value CATE, we keep most part of the formulation unchanged, including the base models for estimating the outcome (\ref{eqn:xlearner-mu}), propensity score model (\ref{eqn:xlearner-propensity}), as well as the final CATE estimation equation (\ref{eqn:xlearner-predict-cate}). The main modification happens to the pseudo-effects (\ref{eqn:xlearner-pseudo-effects}).
Instead of constructing the pseudo-effects on the original outcome variable $Y$, the modified \textit{X-Learner} formulates the net value pseudo-effects as:
    \begin{equation}
    \label{eqn:nvcate-xlearner-pseudo-effects}
    \begin{split}
    \tilde{D}_i^{t_j, t_0} (x^{t_j}_i, Y^{t_j}_i) = \\
    (v - s_{t_j}) Y^{t_j}_i - (v - s_{t_0}) \mu_{t_0}(x^{t_j}_i) - (c_{t_j} - c_{t_0}) \\
    \tilde{D}_i^{t_j, t_0} (x^{t_0}_i, Y^{t_0}_i) = \\
    (v - s_{t_j}) \mu_{t_j}(x^{t_0}_i) - (v - s_{t_0}) Y^{t_0}_i - (c_{t_j} - c_{t_0}) 
    \end{split}
    \end{equation}

where notation of the form $x^{t_j}_i$ indicates we have used the observations form the $j$-th treatment group.
    
A standard regression model can be fit on the net value pseudo-effects, and the corresponding estimator is denoted as $\hat{\tau}_{t_j}(x)$ for treatment $t_j$. The rest of the algorithm follows the same as the \textit{X-Learner} for multiple treatments. 

\subsubsection{Extending R-Learner for Net Value CATE}
To use \textit{R-Learner} for net value CATE estimation, most of the model structure and base model can be re-used. The component that needs to be modified is the optimization equation for CATE.
For \textit{R-Learner}, the standard CATE estimate is specified in equation (\ref{eqn:rlearner-tau}), the CATE estimate on net value can be derived as: 
\begin{equation}
\label{eqn:rlearner-tau}
\begin{split}
\hat{\tau}(\cdot) = argmin_{\tau}\bigg\{ \frac{1}{n} \sum\limits_{i=1}^n \big(\big((v - s_i)Y_i - (v - \bar{s}) \hat{m}^{(-i)}(X_i) - (c_i - \bar{c})\big) \\
- \big(W_i - \hat{e}^{(-i)}(X_i)\big)\tau(X_i)\big)^2 + \Lambda_n(\tau(\cdot))\bigg\}
\end{split}
\end{equation}
where $s_i$ and $c_i$ are the cost based on the observed treatment on user $i$, $\bar{s}$ and $\bar{c}$ are the sample average of cost in the training dataset, $\hat{e}^{(-i)}(X_i)$ and $\hat{m}^{(-i)}(X_i)$ are the same estimators as the regular \textit{R-Learner}.

\section{Empirical Evaluation} \label{empirical_evaluation}
In this section, the proposed models are evaluated empirically with both synthetic data and real data examples. This study also includes models that have been extended to the multiple treatment groups setting previously \cite{Rzepakowski2012-br, Zhao2017-kg}. 


\subsection{Synthetic Data Example}

\begin{figure*}[h]
\includegraphics[width=1\textwidth]{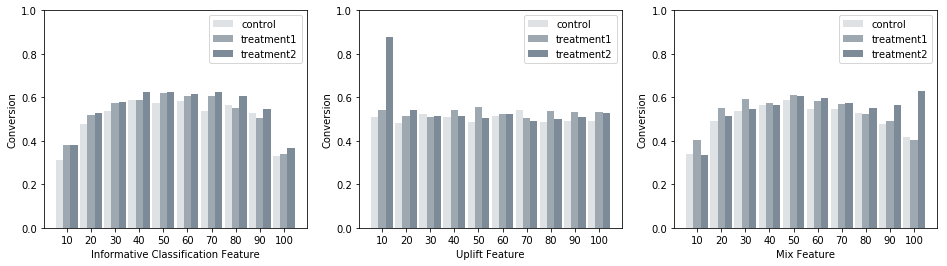}
\caption{An illustration of the relationship between the response variable and different types of features: an informative classification feature influencing the baseline conversion for all treatments, an uplift feature influencing the treatment effect, and a mix feature influencing both the baseline and the treatment effect.}
\label{fig:data_gen}
\end{figure*}

The synthetic data for uplift modeling is generated based on a modified version of the classification data generation algorithm implemented in scikit-learn \cite{scikit-learn} Python library based on the algorithm designed for generating the Madelon dataset in \cite{guyon2005result}. There are different types of features in the uplift modeling data generation: (1) the informative classification feature influences the conversion probability for each unit, but does not affect treatment effect; (2) the uplift feature influences the treatment effect; (3) the mix feature is a random linear combination of the uplift feature and informative classification feature; (4) the irrelevant feature is independent of conversion rate and treatment effect. An example illustration of the relation between these features and the conversion can be found in Figure \ref{fig:data_gen}.

The steps to generate the synthetic data are as follows:
\begin{itemize}
\item Create $n$ units for each treatment group.
\item For the $i$th unit, generate the conversion $Y_i$ based on the informative classification features with a base conversion probability $p_0$, using the standard classification data set generation algorithm \cite{guyon2005result}. At this step, each treatment has the same expected mean of conversion. 
\item For each user in the treatment group $t_j$, generate the potential change $Y'_i$ based on the uplift features with a uplift probability $\delta_j$, using the standard classification data set generation algorithm \cite{guyon2005result}. And then update the the conversion label: $Y_i = \min(Y_i+Y'_i,1)$. The uplift features are different for different treatment groups in this study.
\item Create the mix feature by a random linear combination of a randomly selected uplift feature and a randomly selected informative classification feature. The linear coefficient is drawn from a standard uniform distribution of $[-1,1]$. 
\item Generate the irrelevant features by randomly sampling from the standard normal distribution $N(0,1)$.
\end{itemize}

\subsubsection{Standard Two-Arm Trial}
In this section, we compare the performance of the proposed models in a standard two-arm trial setting. This setting corresponds to a typical A/B test and it is the kind of setting on which the majority of uplift modeling literature has focused. Table \ref{tab:sim_spec} presents the parameters for the experiment.

\begin {table}
\begin{center}
\caption {Parameters of the synthetic data experiments. Most of the experiment groups had both positive and negative lifts depending on the features. We used same random forest tuning parameters across experiments.}
\label{tab:sim_spec}
{\small
\begin{tabular}{ | l | c | c | c |}
\hline
Parameter & Two-arm trial & Four-arm trial & No-control \\
\hline
Sample size & 50000 & 50000 & 30000 \\
Control & Yes & Yes & No \\
Control lift (neg. lift) & 0\% (10\%) & 0\% (0\%) & NA \\
Treat. 1 lift (neg. lift) & 25\% (12.5\%) & 1\% (0.5\%) & 1\% (0.5\%)  \\
Treat. 2 lift (neg. lift) & NA & 2\ (1\%) & 2\ (1\%)  \\
Treat. 3 lift (neg. lift) & NA & 1\% (0\%) & 1\% (0\%) \\
Number of trees & 100 & 100 & 100\\
Max features & 8 & 8 & 8\\
Max depth & 10 & 10 & 10\\
Min samples per leaf & 100 & 100 & 100\\
\hline
\end{tabular}
}
\end{center}
\end{table}

The results of the experiment are shown in Table \ref{tab:synth_results}. Our evaluation metric is based on the Area Under the Uplift Curve (AUUC). \cite{Soltys2015-be, Zhao2017-kg, Rzepakowski2012-br, Gutierrez2016-co}  This metric is calculated by sorting the observations in the testing set to 100 bins from the highest predicted uplift to the lowest. Because the treatment assignment is randomized, we have an approximately equal amount of treatment and control observation in each bin, which allows us to calculate the average treatment effect within each bin. We then calculate the average difference between the treatment and the control if only the highest $p$ bins in the treatment are treated with the treatment condition. The best performing method according to this criterion is the one with the largest AUUC.

\begin{figure}[h]
\includegraphics[width=0.45\textwidth]{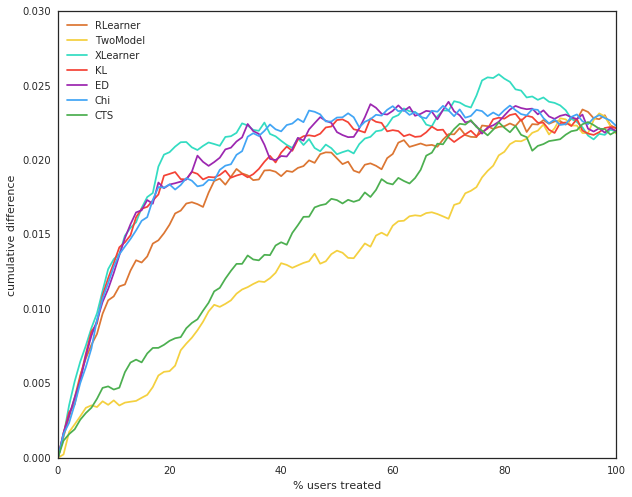}
\caption{The uplift curves for various algorithms under the two-arm setting. The \textit{R-Learner} and \textit{X-Learner} both perform better than the other algorithms in the comparison.}
\label{fig:two_treatment_uplift_curve}
\end{figure}

Figure \ref{fig:two_treatment_uplift_curve} presents the uplift curves for various algorithms in the two-arm experiment. 

\begin {table}
\begin{center}
\caption {Results of the synthetic data experiments. The ``Time'' columns presents the computing time in seconds.}
\label{tab:synth_results}
{\small
\begin{tabular}{ | l | c | c | c | c |  }
\hline
\multicolumn{1}{|c}{} & \multicolumn{2}{c}{Two-arm trial} & \multicolumn{2}{c|}{Four-arm trial} \\
\hline
Model & Time & AUUC & Time & AUUC  \\
\hline         
\textit{R-Learner} & 85.11s & 0.0185 & 111.17s & 0.0310 \\
\textit{X-Learner} & 84.03s & 0.0205 & 125.68s & 0.0305 \\
\textit{Two Model} & 35.60s & 0.0135 & 54.86s & 0.0283 \\
\textit{KL} & 1664.90s & 0.0195 & 1460.01s & 0.0331 \\
\textit{\textit{ED}} & 1645.73s & 0.0201 & 1423.55s & 0.0319 \\
\textit{Chi} & 1598.44s & 0.0201 & 1408.22s & 0.0329 \\
\textit{CTS} & 1577.24s & 0.0153 & 1254.25s & 0.0270 \\
\hline 
\end{tabular}
}
\end{center}
\end{table}

\subsubsection{Multiple Treatment Groups with a Control}
In this experiment, we move to the multiple treatment group setting. Here, we compare three treatment groups with different kinds of treatment effects and informative features against a control group that does not receive any treatment. The parameters for this experiment, which represents a situation that is very common in industry applications, are presented in Table \ref{tab:sim_spec}.

\begin{figure}[h]
\includegraphics[width=0.45\textwidth]{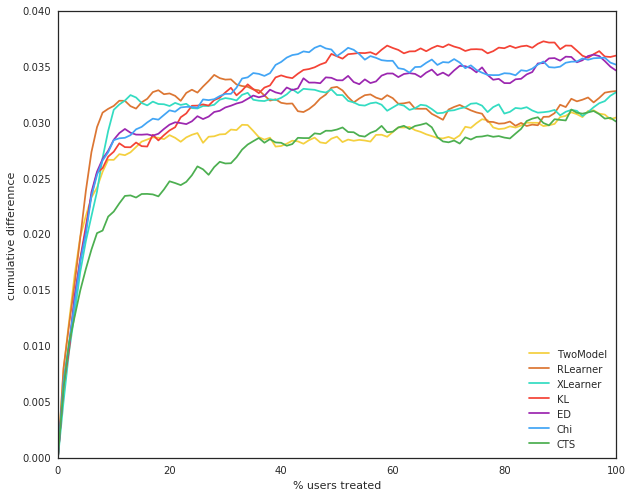}
\caption{The uplift curve for various algorithms in the setting with three treatment groups and a control. The multi-treatment versions of \textit{R-Learner} and \textit{X-Learner} developed here provide better performance than the other algorithms proposed for the multi-treatment case.}
\label{fig:three_treatment_uplift_curve}
\end{figure}

Uplift modeling in this kind of setting answers the question of whether an individual should receive a treatment and, if so, which treatment they should receive. As before, the AUUC is calculated by sorting the observations in the test set according to their predicted uplift and then calculating the average uplift that would be captured if the highest $p$ percent of these observations were to receive the recommended treatment. However, this time the average treatment effect calculation within a bin involves only the control observations and those observations that happened to be in the treatment group in which they are predicted to have the largest treatment effect. This allows us to compare the prediction of the model to actually observed treatment effects. Figure \ref{fig:three_treatment_uplift_curve} presents the uplift curves for the models under comparison and Table \ref{tab:synth_results} summarizes the results. Unlike the two-arm scenario, the multiple-arm uplift curve can ends with different cumulative difference when $100 \%$ users are treated. In the two-arm case, the ending difference reflects the ATE for the treatment compared with the control, while in the multi-arm case, the ending difference reflects the ATE for the optimal personalized treatment the uplift model could find compared with control. 

\subsubsection{Multiple Treatments Without a Control}
\begin{figure}[h]
\includegraphics[width=0.45\textwidth]{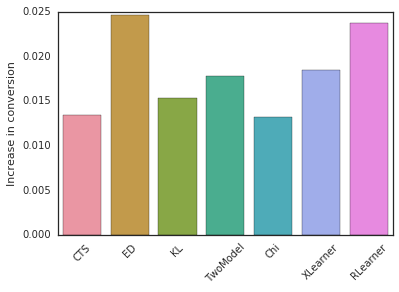}
\caption{The Results of the No-control (Three-Arm) Optimisation Task. The bars correspond to the difference in the outcome between those observations in the test sample who were in the recommended treatment group and those observations who were not in the recommended treatment group. The treatment group allocation achieved by the \textit{X-Learner} and the \textit{R-Learner} is comparable or better than the other approaches, although we also see strong performance by the \textit{ED} algorithm.}
\label{fig:three_treatment_optimization}
\end{figure}

Another typical experimental setting is one in which we have a number of different treatment groups but no control group that would receive no treatment. In this setting, our goal is to determine which one out of the different treatment groups is the best one for a given individual in terms of the predicted uplift. We compared the performance of the algorithms using a dataset with three treatment groups, using the simulation parameters described in Table \ref{tab:sim_spec}.

To determine the treatment group recommended by each algorithm, we ran pairwise comparisons between the treatments and selected a winner in each comparison based on the predicted uplift. The recommended treatment group for an individual was decided by majority voting. If the voting resulted in a tie, we assigned one of the recommended treatment groups randomly. As the outcome metric, we looked at the difference in outcome between those who were in the recommended treatment group and those who were not. Figure \ref{fig:three_treatment_optimization} presents the results of the experiment.

\subsubsection{Net Value Optimization}

\begin{figure*}[h]
\centering
\includegraphics[width=1\textwidth]{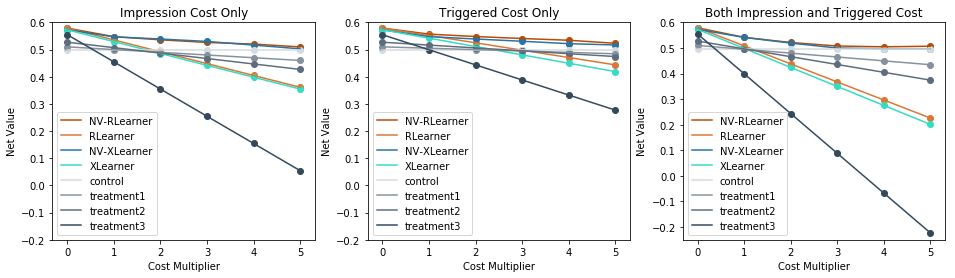}
\caption{Net Value Optimization with Varying Cost. In the net value optimization task, the net value meta-learner models show better results than the standard meta-learner models and the fixed treatment groups.}
\label{fig:syn_nv}
\end{figure*}

To evaluate the net value optimization models, the conversion value and treatment cost are added in the simulated dataset. The conversion value is assumed to be constant $v=1$. The cost is varied in different simulation trials: denoting $c_m$ as a cost multiplier taking values in $\{0,1,2,3,4,5\}$, the impression cost is $s_{t0} = 0, s_{t1} = 0.01 c_m, s_{t2} = 0.02 c_m, s_{t3} = 0.1 c_m$, and the triggered cost is $s_{t0} = 0, s_{t1} = 0.01 c_m, s_{t2} = 0.1 c_m, s_{t3} = 0.5 c_m$. To test the algorithms with different scenarios, the trials are designed to have: (i) impression cost only, (ii) triggered cost only, and (iii) both impression and triggered cost. In the trials with an impression cost only, all costs are assumed to be $0$, and the triggered cost is varying with $c_m$; and vice versa for the trials with a triggered cost only. In the trials with both impression and triggered costs, both cost variables are varying with $c_m$.

The results are summarized in Figure \ref{fig:syn_nv}. Among the different experiment groups, Treatment $3$ has the highest conversion but also the highest cost, which leads to the lowest net value if the treatment is applied to all users. Control group, again, has no cost but the conversion rate is the lowest. The modeling goal is to personalize the experience in order to achieve the optimal net value. The uplift models (Standard \textit{R-Learner} and Standard \textit{X-Learner}) optimize the conversion rate without considering the cost, which leads to a sub-optimal result in terms of net value. The modified uplift models (Net Value \textit{R-Learner} and Net Value \textit{X-Learner}) achieve net values that overperform the control and any single treatment group, as well as the standard uplift models. This results shows that the net value optimization framework is successful in identifying the net value CATE at the individual level and making the tradeoff between value and cost.

\subsection{Real Data Example}
In this example, the proposed uplift models are applied to a real business use case. The dataset is collected from an online experiment of a promotion campaign. In this experiment, there are one control group and two treatment groups. In the control group, the users receive no communications. In the treatment groups, different types of promotions are sent to the users. The goal of the promotion is to encourage users to start using a new product. While running a randomized experiment helps the team identify the best performing promotion and quantify its impact, applying an uplift model enables the team to personalize the targeting to maximize the net value given the budget. The value and cost numbers have been re-scaled to hide sensitive business information, but the evaluation statistics are representative of model performance. 

The cost structure in the dataset is as follows. The control is the default experience and does not have any cost associated with it, treatment $1$ is assigned an impression cost of $0.01$ and a triggered cost of $0.2$, and treatment $2$ is assigned an impression cost of $0.01$ and a triggered cost of $0.3$. The triggered cost corresponds to the promotion amount when the user redeems the voucher, and the impression cost is the cost associated with the communication channel. Although the direct cost for the communication channel is minimal, the hidden cost is considerable: sending too many emails to a user can lead to unsubscribing. Such a hidden cost is quantified and included in the impression cost. The conversion value for each individual user is assumed to be $1$.

The data is split into training data ($600$K observations) and testing data ($800$K observations). The models are trained on the training data, and applied to the testing data to report the performance. There are $139$ features in the data. The hyper-parameters used for the random forests models are: $50$ trees, $10$ features selected at each split, $10$ as maximum depth, and $100$ as minimum sample in each leaf.

\begin{figure}
\centering
\includegraphics[width=0.45\textwidth]{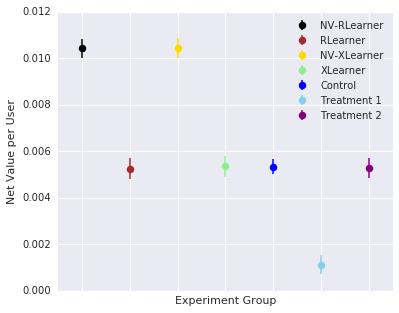}
\caption{Average Net Value Per User with $95\%$ Confidence Interval in Real Data Example. Both net value models NV-Rlearner and NV-Xlearner perform significantly better than other models and original experiment groups.}
\label{fig:real_data_nv}
\end{figure}

Figure \ref{fig:real_data_nv} shows the results from different models on the testing data. Both net value optimization meta-learners perform significantly better in terms of the average net value than other models and original experiment groups. In contrast, neither of the standard meta learners is able to improve the net value, since they do not take the costs into account. It is worth noting that there is no guarantee the average conversion rate is higher in the net value optimization groups, since the models make a trade-off between the likelihood of incremental conversion and the associated cost. As an illustration of this trade-off, consider the conversion rates in the treatment groups, which are $0.53\%$ in Control, $1.39\%$ in Treatment 1 and $2.18\%$ in Treatment 2. While the conversion in Treatment 1 is much better than in Control, the net value is worse when the relevant costs are taken into account. One typical situation in which the net value optimization algorithms can reduce the average conversion while increasing the expected net value is when the incremental conversion value (as the incremental conversion probability times the conversion value) of the promotion is lower than the known cost.

\section{Platform Implementation} \label{implementation}

\begin{figure}[h]
\includegraphics[width=0.47\textwidth]{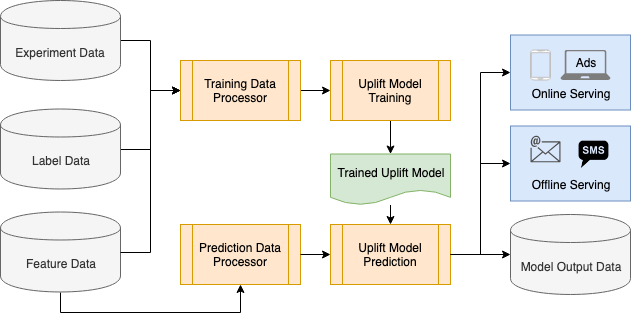}
\caption{Flowchart for Uplift Modeling Implementation for Online and Offline Use Cases}
\label{fig:uplift_flow}
\end{figure}

To meet the increasing demand for applying uplift modeling in different business scenarios at Uber, the uplift models are being implemented at scale in a machine learning platform for a broad set of use cases. Figure \ref{fig:uplift_flow} shows the system design for the platform implementation. The training data processor combines treatment group tags, outcome labels and user features into a consumable data frame for model training. The model training module fits multiple uplift models and selects the best performing model based on an evaluation metric (such as the AUUC score) using cross-validation. The trained model is then stored in a model file to be picked up by a model prediction module. The model prediction module makes predictions on users who are considered as the target cohort for a new campaign. The generated user-level uplift scores can be pushed to an online service (such as in-product notification and recommendation, advertisement, etc.), offline service (such as customer relationship management platform for email, SMS, etc.), or a data store. 

The system takes user inputs and configurations for selecting the target metric ($Y$) and features, identifying the corresponding experiment, defining a cohort filter if a segment of users is of interest, configuring the model hyperparameters (including the cost and value assignment), and determining the targeting cohort cut-point (e.g. selecting top $50\%$ of users with the highest uplift score for final targeting). The system outputs can be produced through different procedures: the online service can call the model predictor as a service and get the predicted score in real time, or the offline service can schedule the model predictor to run batch jobs on a designed cadence and push the batch results to a destination data base.

\section{Conclusion} \label{conclusion}
Multiple treatment groups are commonly seen in practice when there are multiple channels, designs, content, or strategies to provide a service to users. In this paper, we extended the existing framework of meta-learners to the multiple treatment groups case. In addition, we proposed net value optimization models to take the value of conversion as well as the varying costs of treatments into account. We evaluated these models empirically in a number of scenarios against other models that have been proposed for the multiple treatment case. In the empirical study, the proposed models showed comparative advantage in optimization measures and computation speed. As uplift modeling is being implemented on a platform, one direction for future work is to discuss the practical challenges and lessons from large-scale applications in multiple areas. For example, examining and improving the quality of training data can be critical for model performance. Another area for future work is to extend the proposed uplift models to regression problems, where the incremental value in a continuous metric is of interest for optimization. 

\section*{Acknowledgment}
We would like to thank Fran Bell for providing insightful feedback and comments on this paper, Neha Gupta and Candice Hogan for the support on the project development, Hugh Williams for general guidance on the uplift modeling methodology, James Lee and Shuyang Du for discussions on implementing uplift models in production, and Yunhan Xu, Mert Bay, Yuchen Luo, Lance Mack and Aiste Juknaite for collaboration on use cases. We would also like to thank a number of anonymous reviewers for helpful comments on earlier versions of this paper.

\bibliographystyle{./IEEEtran}
\bibliography{uplift_references}

\end{document}